\documentclass[10pt,twocolumn,letterpaper]{article}

\usepackage{cvpr}
\usepackage{times}
\usepackage{epsfig}
\usepackage{amsmath}
\usepackage{amssymb}
\usepackage{graphicx}
\usepackage{booktabs}
\usepackage[pagebackref=true,breaklinks=true,colorlinks,bookmarks=false]{hyperref}

\begin{document}

\title{Measuring Domain Shifts using Deep Learning Remote Photoplethysmography Model Similarity}

\author{Nathan Vance and Patrick Flynn\\
University of Notre Dame\\
{\tt\small \{nvance1,flynn\}@nd.edu}
}

\maketitle

\begin{abstract}
    Domain shift differences between training data for deep learning models and the deployment context can result in severe performance issues for models which fail to generalize. We study the domain shift problem under the context of remote photoplethysmography (rPPG), a technique for video-based heart rate inference. We propose metrics based on model similarity which may be used as a measure of domain shift, and we demonstrate high correlation between these metrics and empirical performance. One of the proposed metrics with viable correlations, \texttt{DS-diff}, does not assume access to the ground truth of the target domain, \ie it may be applied to in-the-wild data. To that end, we investigate a model selection problem in which ground truth results for the evaluation domain is not known, demonstrating a 13.9\% performance improvement over the average case baseline.
\end{abstract}
\section{Introduction}

Remote Photoplethysmography (rPPG) is a technique for inferring a subject's blood volume pulse (BVP) waveform from video data~\cite{DeHaan2013,wang2016algorithmic,yu2019remote,liu2020multi}. Approaches for rPPG are varied, comprising both classical techniques~\cite{DeHaan2013,wang2016algorithmic} and deep learning based methods~\cite{yu2019remote,liu2020multi}.

A model undergoes a domain shift when it is transferred from its training domain to an evaluation domain. This is a serious problem in rPPG; if improperly addressed, a model which obtains state of the art results under a domain closely related to its training domain may fail catastrophically when deployed for an in-the-wild application. As an example from a different application, in \cite{koenecke2020racial} significant racial bias was found in the form of a performance gap between white and black users of machine learning based automated speech recognition (ASR) services deployed by major companies. The authors trace the performance gap to deficiencies in the acoustic models, and particularly to ``insufficient audio data from black speakers when training the models.'' In other words, the domain shift for these acoustic models between the training dataset and the more race-balanced real world domain revealed an algorithmic bias, which has resulted in harsh criticism leveled at the entire ASR industry~\cite{mengesha2021don}.

This problem is not unique to ASR. Tang \etal found that deep learning rPPG models trained on datasets such as PURE and UBFC-rPPG, which contain only subjects with Fitzpatrick skin tones 2 and 3, underperform on subjects with darker skin tones~\cite{tang2023mmpd}. Skin tone is only one factor resulting in a domain shift --- other often cited factors include motion~\cite{DeHaan2013}, illumination~\cite{tang2023mmpd}, and heart rate range~\cite{vance2023generalization}.

\begin{figure}
    \centering
    \includegraphics[width=\linewidth]{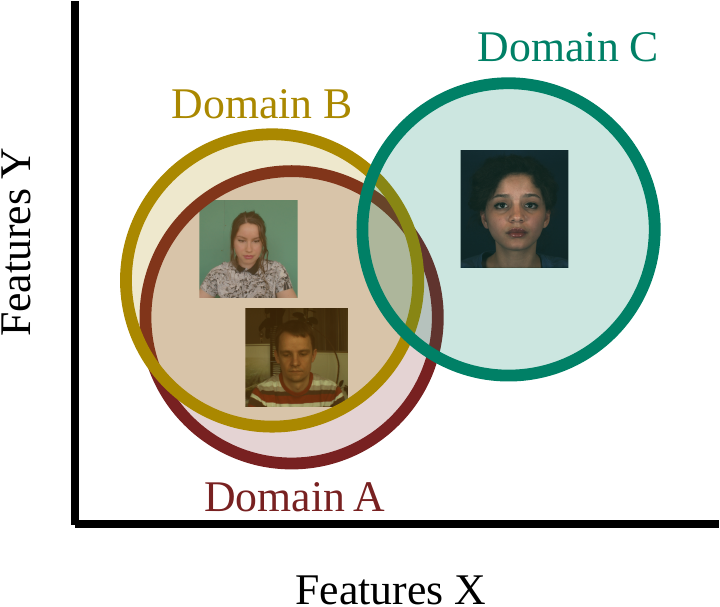}
    \caption{Illustration of a domain shift in feature space. A mild domain shift exists between Domains A and B, and a severe domain shift exists between Domains A and C, and B and C. We propose systems based on model similarity for measuring the domain shifts.}
    \label{fig:domainshift-diagram}
\end{figure}

A diagram depicting a domain shift is shown in Figure \ref{fig:domainshift-diagram}. In feature space, the domain shift is realized as the difference in representations of the data making up the domain. If two datasets are very similar, such as PURE and UBFC-rPPG containing only lightly skinned subjects, then their features are likely to be closely aligned as with Domains A and B in the diagram. However, a dataset containing darker skinned subjects such as BP4D+~\cite{zhang2016multimodal} could constitute a severe domain shift, resulting in features that differ from the other domains as with Domain C in the diagram. As the diagram implies, our approach to domain shifts is to represent them as a feature-level phenomenon; \ie, while the root cause of a domain shift may be intrinsic to properties of the data, the manifestation is in terms of the features extracted from the data, \eg by deep learning models.

Building on the findings from \cite{vance2024cka}, we investigated the use of Centered Kernel Alignment (CKA) as a tool for the measurement of domain shifts. In particular, CKA measures the similarity of model activations (\ie features) between layers of models as they are evaluated over data~\cite{kornblith2019similarity}. We propose techniques based on CKA as a means to measure the distance between domains. The resulting tooling benefits both our scientific understanding of how datasets relate one to another, and also provides a CKA-based utility for detecting and mitigating domain shifts at inference time, even with unknown data~\cite{stacke2020measuring,elsahar2019annotate,chakrabarty2023simple}. As far as the authors are aware, we are the first to explore the domain shift measurement problem under the context of rPPG, and the first to propose and evaluate CKA-based domain shift measurement metrics.

The contributions of this work are as follows:

\begin{itemize}
    \item We propose metrics based on model similarity for measuring the domain shift between datasets.
    \item We conduct experiments demonstrating high correlation between these metrics and empirical performance over rPPG datasets.
    \item We provide a model selection application for one of the proposed metrics, \texttt{DS-diff}, evaluating its performance against worst case, best case, and average case baselines.
\end{itemize}
\section{Related Work}

\begin{figure*}
    \centering
    \subfloat[DDPM]{\includegraphics[height=.2\textwidth]{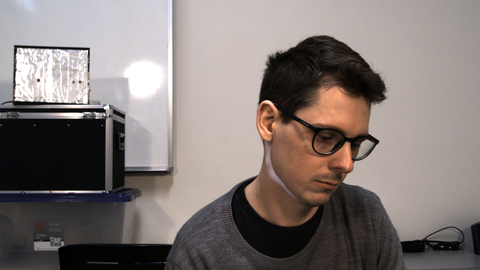}\label{fig:ds:ddpm}}
    \subfloat[UBFC-rPPG]{\includegraphics[height=.2\textwidth]{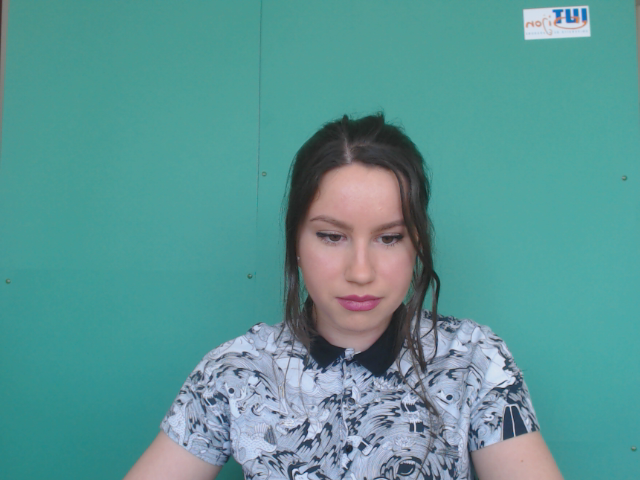}\label{fig:ds:ubfc}} \\
    \subfloat[PURE]{\includegraphics[height=.2\textwidth]{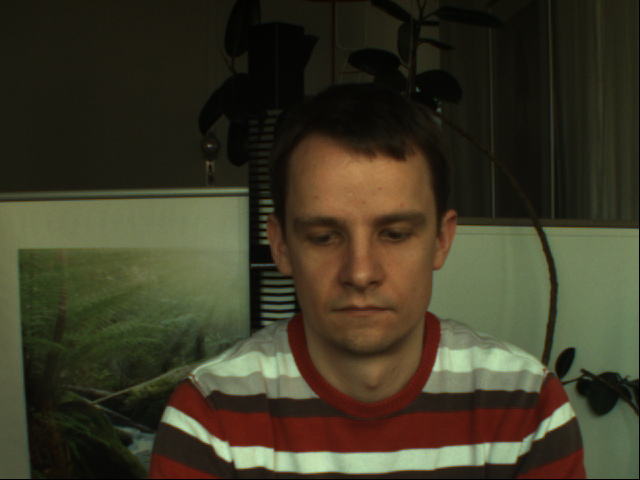}\label{fig:ds:pure}}
    \subfloat[BP4D+]{\includegraphics[height=.2\textwidth]{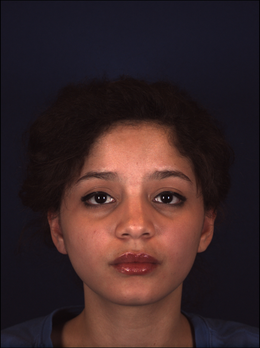}\label{fig:ds:bp4d}}
    \subfloat[MSPM]{\includegraphics[height=.2\textwidth]{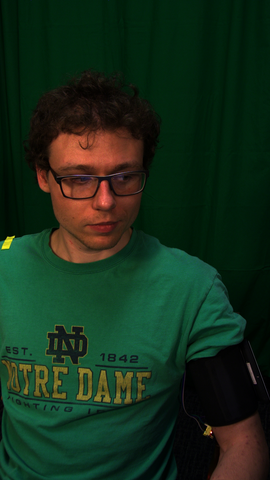}\label{fig:ds:mspm}}
    \caption{Sample images from each of the utilized datasets. BP4D+ (Figure \ref{fig:ds:bp4d}) and MSPM (Figure \ref{fig:ds:mspm}) are further subdivided to generate 21 datasets total.}
    \label{fig:ds}
\end{figure*}

This work studies the measurement of domain shifts using model similarity based metrics. We therefore survey advances in the study of domain shifts in Section \ref{sec:related:domainshift}, as well as techniques used for model similarity measurements in Section \ref{sec:related:netsim}.

\subsection{Domain Shift Effects and Mitigation} \label{sec:related:domainshift}

When a model is transferred from its training domain to a testing domain it undergoes a \textit{domain shift}. The shift may be very slight such as when the testing domain is a randomly selected withheld test set from the training dataset, or it can be severe such as when a model is trained on laboratory data and deployed for ``in the wild'' use.

In 2019, Stacke \etal investigated domain shifts in the context of histopathology~\cite{stacke2020measuring}. They utilize feature map differences obtained by the same model on different domains to calculate the representation shift, a metric related to the domain shift. In particular, they demonstrate a clear negative correlation between representation shift and classification accuracy. Similarly, Elsahar and Gall{\'e} investigate a suite of domain shift measurements for binary classifiers, applying these to NLP problems~\cite{elsahar2019annotate}. Chakrabarty \etal develop a simple domain shift detection metric based on batch-averaged features for selective application of test-time adaptation methods~\cite{chakrabarty2023simple}. We base our domain shift metrics on CKA~\cite{kornblith2019similarity} as it is a mature measure of model activation similarity suitable for use as a building block in domain shift measurements.

Techniques have been developed to mitigate domain shifts. A common technique is to employ adversarial learning to align model representations across domains as in \cite{sankaranarayanan2018learning,luo2019taking,wang2023hierarchical,du2023dual}. Other techniques utilize unsupervised adaptive learning to fine-tune pretrained models to new domains~\cite{zhang2021adaptive,kanakasabapathy2021adaptive}. Rather than explore interventions which boost the performance of a given model in light of a domain shift, in Section \ref{sec:modelselect} we apply domain shift measurement techniques to inform model selection.

\subsection{Neural Network Similarity} \label{sec:related:netsim}

The goal of neural network similarity is to compare neural network behavior, quantifying the functional similarity of networks while remaining invariant to incidental differences, \eg those which arise simply from differing random initializations. Rather than probe neural weights directly, the methods surveyed in this section perform this comparison over network representations in the form of activations as the networks process data.

Early pioneers in neural network similarity include Laakso and Cottrell who, in 2000, proposed a network similarity technique based on the distance between neural network activations~\cite{laakso2000content}. Raghu \etal developed Singular Vector Canonical Correlation Analysis (SVCCA), a method that permits neural network similarity comparisons between different layers, including across architectures~\cite{raghu2017svcca}. Morcos \etal extend SVCCA to better differentiate between signal and noise, calling their technique Projection Weighted CCA (PWCCA)~\cite{morcos2018insights}. Kornblith \etal propose Centered Kernel Alignment (CKA)~\cite{kornblith2019similarity}, a technique based on a normalized Hilbert-Schmidt Independence Criterion~\cite{gretton2005measuring}, which is shown to be more reliable than CCA-based methods on models trained from different initializations. We make use of the torch\_cka implementation of CKA~\cite{subramanian2021torch_cka} as a building block for the methods proposed in this work. Cui \etal propose Deconfounded CKA, an adjustment to CKA promoting robustness against domain shifts~\cite{cui2022deconfounded}. We find this modification to be antithetical to our intended use case of CKA, rather desiring a neural network similarity measure that is sensitive to domain shifts.
\section{Datasets} \label{sec:domainshift_datasets}

\begin{table*}
    \centering
    \begin{tabular}{l|ccc}
\toprule
Dataset          & Time (s)             & Avg HR (BPM)              & Avg HR stddev (BPM)      \\
\midrule
BP4D-1           & 28.883 $\pm$ 2.424   & 104.226 $\pm$ 2.960 & 4.773 $\pm$ 0.835 \\
BP4D-2           & 27.142 $\pm$ 1.210   & 96.784 $\pm$ 2.405  & 3.328 $\pm$ 0.408 \\
BP4D-3           & 73.343 $\pm$ 0.115   & 94.857 $\pm$ 2.403  & 3.533 $\pm$ 0.560 \\
BP4D-4           & 15.909 $\pm$ 0.274   & 102.656 $\pm$ 3.497 & 7.084 $\pm$ 1.723 \\
BP4D-5           & 28.016 $\pm$ 1.301   & 98.426 $\pm$ 2.646  & 2.838 $\pm$ 0.389 \\
BP4D-6           & 49.465 $\pm$ 3.022   & 102.582 $\pm$ 2.668 & 6.323 $\pm$ 0.797 \\
BP4D-7           & 33.423 $\pm$ 1.399   & 97.581 $\pm$ 2.723  & 3.641 $\pm$ 0.500 \\
BP4D-8           & 50.378 $\pm$ 2.705   & 104.589 $\pm$ 3.100 & 5.429 $\pm$ 0.932 \\
BP4D-9           & 108.994 $\pm$ 5.491  & 99.365 $\pm$ 2.615  & 8.347 $\pm$ 0.984 \\
BP4D-10          & 15.459 $\pm$ 1.303   & 99.397 $\pm$ 3.233  & 1.497 $\pm$ 0.526 \\
DDPM             & 632.096 $\pm$ 31.532 & 98.351 $\pm$ 4.559  & 8.903 $\pm$ 1.525 \\
MSPM-adversarial & 120.133 $\pm$ 0.000  & 77.698 $\pm$ 2.481  & 4.198 $\pm$ 0.873 \\
MSPM-game        & 145.925 $\pm$ 4.070  & 79.358 $\pm$ 2.582  & 5.298 $\pm$ 0.956 \\
MSPM-hold        & 74.272 $\pm$ 3.418   & 83.121 $\pm$ 2.896  & 8.132 $\pm$ 0.680 \\
MSPM-relax1      & 138.729 $\pm$ 1.880  & 81.085 $\pm$ 2.667  & 5.221 $\pm$ 0.826 \\
MSPM-relax2      & 77.001 $\pm$ 1.815   & 78.621 $\pm$ 2.455  & 4.905 $\pm$ 0.509 \\
MSPM-respiration & 178.957 $\pm$ 0.096  & 82.798 $\pm$ 2.799  & 4.676 $\pm$ 0.458 \\
MSPM-videoBW     & 66.019 $\pm$ 0.148   & 75.688 $\pm$ 3.166  & 3.259 $\pm$ 0.317 \\
MSPM-videoRGB    & 52.155 $\pm$ 0.044   & 76.701 $\pm$ 2.598  & 3.274 $\pm$ 0.922 \\
PURE             & 68.307 $\pm$ 1.502   & 69.200 $\pm$ 6.026  & 2.340 $\pm$ 0.352 \\
UBFC-rPPG        & 64.964 $\pm$ 1.516   & 100.801 $\pm$ 5.056 & 4.994 $\pm$ 0.890 \\
\bottomrule
\end{tabular}
    \caption{Summary statistics of datasets and sub-datasets used in domain shift experiments.}
    \label{tab:domainshift-summarystats}
\end{table*}

As the aim of these experiments is to study domain shifts, five different datasets are utilized, two of which are divided into subsets yielding 21 domains in total. In particular:  DDPM~\cite{vance2022deception}, UBFC-rPPG~\cite{Bobbia2019}, and PURE~\cite{Stricker2014} are utilized as full datasets, along with 10 subsets of BP4D+~\cite{zhang2016multimodal} and 8 subsets of MSPM~\cite{speth2024mspm}. Sample images taken from each of these datasets are displayed in Figure \ref{fig:ds}. The subsets generated from BP4D+ and MSPM are as follows:

\begin{itemize}
    \item \textbf{BP4D+}: We partition the dataset by activity into 10 sub-datasets BP4D-X, where X is the activity number as reported in~\cite{zhang2016multimodal}. In particular:
    \begin{enumerate}
        \item Listen to a funny joke
        \item Watch a 3D avatar of participant
        \item Watch a video clip of a 911 emergency phone call
        \item Experience a sudden burst of sound
        \item Interview: True or false question
        \item Improvise a silly song
        \item Experience physical threat in a dart game
        \item Cold pressor: Submerge hand into ice water
        \item Interviewer pretends to complain about the subject's poor performance
        \item Experience a smelly odor
    \end{enumerate}
    \item \textbf{MSPM}: We partition the dataset by activity into sub-datasets MSPM-activity, where the activity label used in this work corresponds to the following activity numbers as reported in~\cite{speth2024mspm}:
    \begin{itemize}
        \item 1-2 \textbf{relax1}: First round of relaxation
        \item 3-4 \textbf{respiration}: Guided respiration video
        \item 5-7 \textbf{hold}: Breath hold
        \item 8-9 \textbf{game}: SuperTuxCart video game
        \item 10 \textbf{videoBW}: Watch a clip from It's A Wonderful Life
        \item 11 \textbf{videoRGB}: Watch a clip from Star Wars VI Return of the Jedi
        \item 12 \textbf{adversarial}: Adversarial attack on rPPG
        \item 13-14 \textbf{relax2}: Second round of relaxation
    \end{itemize}
\end{itemize}

We provide summary statistics of the 21 domains in Table \ref{tab:domainshift-summarystats}. These datasets contain a wide spread of durations ranging from just over 15 seconds with BP4D-4 and BP4D-10, up to over 10 minutes with DDPM; average heart rates ranging from under 70 beats per minute (BPM) with PURE up over 100 BPM on UBFC-rPPG and certain subsets of BP4D; and heart rate standard deviations (a measure of heart rate variability) ranging from under 3 BPM on PURE, BP4D-5, and BP4D-10, up above 7 BPM on BP4D-4, BP4D-9, DDPM, and MSPM-hold. While these statistics may not explain the full extent of the domain shifts present in the datasets, heart rate and heart rate variability have been found to be major impediments to generalization~\cite{vance2023generalization}.
\section{Training}

\begin{table}
    \centering
    \begin{tabular}{l|c}
        \toprule
        Dataset          & MAE                \\
        \midrule
        BP4D-1           & 5.456 $\pm$ 2.602  \\
        BP4D-2           & 3.977 $\pm$ 1.887  \\
        BP4D-3           & 2.383 $\pm$ 1.143  \\
        BP4D-4           & 14.372 $\pm$ 2.573 \\
        BP4D-5           & 4.606 $\pm$ 4.276  \\
        BP4D-6           & 5.089 $\pm$ 2.050  \\
        BP4D-7           & 3.839 $\pm$ 1.887  \\
        BP4D-8           & 4.880 $\pm$ 3.176  \\
        BP4D-9           & 4.167 $\pm$ 2.458  \\
        BP4D-10          & 6.028 $\pm$ 4.169  \\
        DDPM             & 3.243 $\pm$ 1.948  \\
        MSPM-adversarial & 10.748 $\pm$ 5.594 \\
        MSPM-game        & 2.737 $\pm$ 1.187  \\
        MSPM-hold        & 3.799 $\pm$ 0.943  \\
        MSPM-relax1      & 2.481 $\pm$ 1.184  \\
        MSPM-relax2      & 5.022 $\pm$ 0.758  \\
        MSPM-respiration & 1.947 $\pm$ 0.879  \\
        MSPM-videoBW     & 2.389 $\pm$ 2.695  \\
        MSPM-videoRGB    & 1.782 $\pm$ 0.587  \\
        PURE             & 2.238 $\pm$ 2.155  \\
        UBFC-rPPG        & 0.610 $\pm$ 0.207  \\
        \bottomrule
    \end{tabular}
    \caption{Training results for 3DCNN-6 models. The 95\% confidence intervals are calculated over a five-folds evaluation.}
    \label{tab:domainshift_training}
\end{table}

We performed training with five-fold cross validation across the 21 domains listed in Section \ref{sec:domainshift_datasets} to generate 
105 models for further analysis. The architecture used in this study is a 6-layer 3DCNN based on PhysNet-3DCNN~\cite{yu2019remote}. This architecture was selected based on findings in \cite{vance2024cka}, in particular, that it exhibits fewer redundancies compared to the standard 10-layer PhysNet-3DCNN while still retaining good empirical performance on the investigated datasets. Accordingly, fewer redundancies in this more compact architecture leads to more informative CKA similarities.

We perform histogram equalization both when training and at inference. When training we apply the augmentations used in \cite{vance2023generalization}, \ie horizontal flip, illumination noise, Gaussian blur, playback speed, and playback modulation. We convert the blood volume pulse (BVP) waveform outputted by the network into heart rates using a short term fourier transform (STFT) with a window size of 10 seconds and a frequency band of 40-180 beats per minute (BPM).

The intra-dataset test set mean absolute error (MAE) from five-fold splits is reported in Table \ref{tab:domainshift_training}. We observe the worst performance in BP4D-4, which is the ``startle activity'' in which subjects experience a sudden burst of sound; it is also a short segment of the dataset at an average of 16 seconds per clip compared to the 51 seconds per clip dataset average. We also observe poor performance for MSPM-adversarial, in which pulsating light is projected onto subjects' faces as a physical attack on rPPG systems.

\begin{figure}
    \centering
    \includegraphics[width=\linewidth]{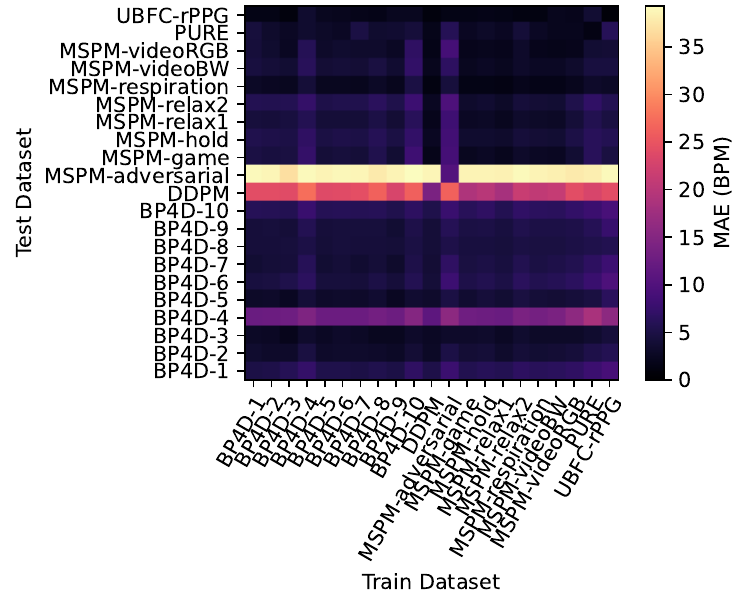}
    \caption{Cross dataset evaluation results reporting MAE for models trained on a source dataset and evaluated on a test dataset. MAE values are averaged across the five folds.}
    \label{fig:domainshift_xds}
\end{figure}

We performed a cross dataset evaluation with results shown in Figure \ref{fig:domainshift_xds}. As was the case in the intra-dataset evaluation, BP4D-4 and MSPM-adversarial obtained poor performance as cross-dataset evaluation targets. DDPM additionally presented a challenging domain shift scenario.

\section{CKA as a Domain Shift Measurement}

We investigated several techniques for utilizing CKA as a measure of domain shift, comparing these techniques to MAE as an empirical measure of domain shift.
MAE is more aptly termed a proxy for domain shift, since a domain shift is only one explanation for why a degradation in empirical performance is observed.

\subsection{Domain Shift Metrics} \label{subsec:domainshift_metrics}

\begin{figure}
    \centering
    \includegraphics[width=\linewidth]{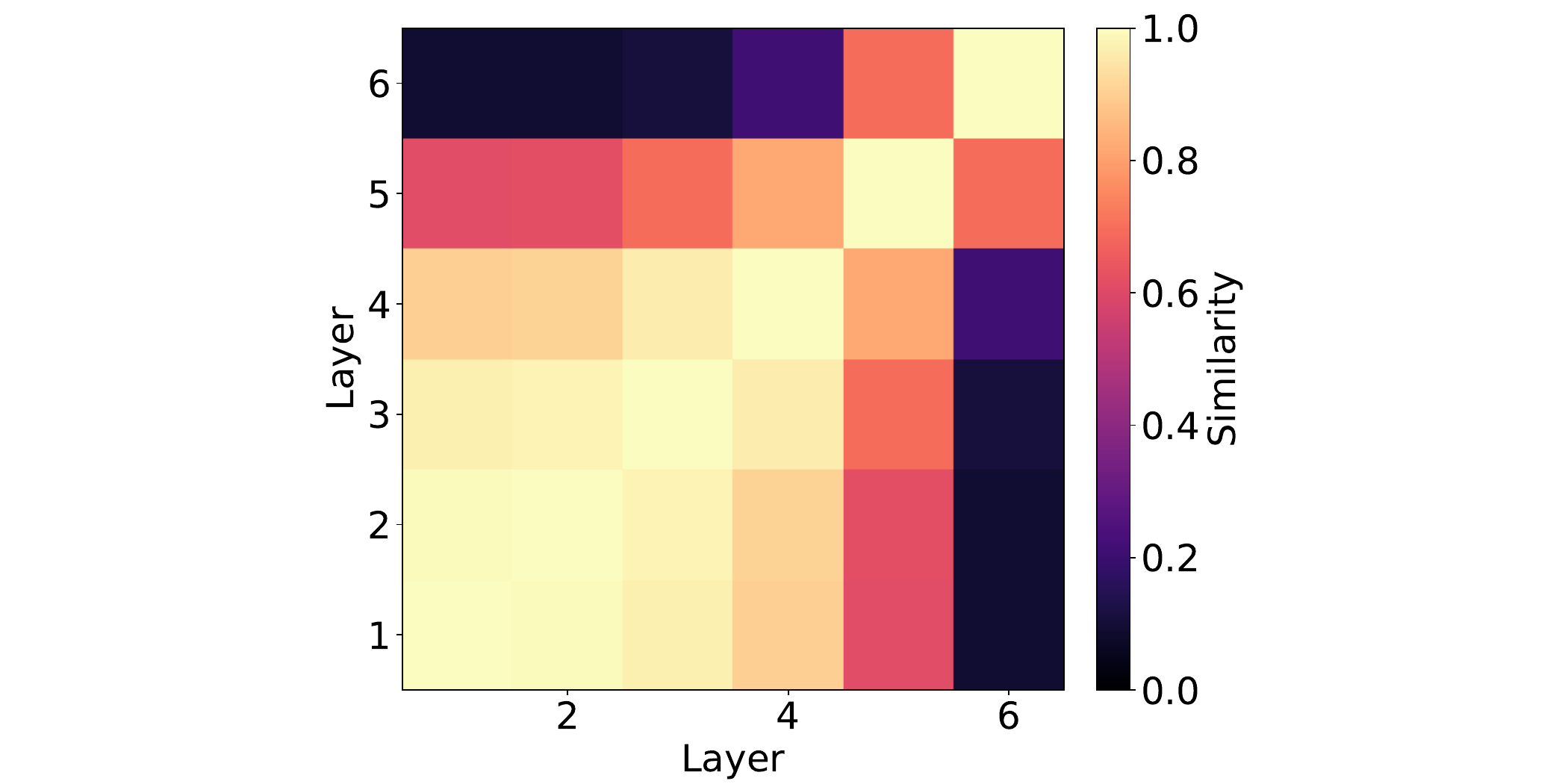}
    \caption{Example CKA map.}
    \label{fig:exampleCKA}
\end{figure}

In this section we describe the CKA-based metrics developed for the measurement of domain shifts, and 
develop notation to precisely describe the common components to these metrics. We define the CKA operation:

\begin{equation}
    map = CKA(m_x, ds_x, m_y, ds_y), 
\end{equation}
\noindent
where $m_x$ and $m_y$ are the models used for performing the CKA analysis along the $x$ and $y$ axes respectively, $ds_x$ and $ds_y$ are the datasets used for performing the CKA analysis along the $x$ and $y$ axes respectively, and $map$ is the resulting CKA map, which is a table whose $x$ and $y$ dimensions are the same length as the number of layers in $m_x$ and $m_y$ respectively. An example CKA map is shown in Figure \ref{fig:exampleCKA}. We utilize the torch\_cka implementation for the experiments in this paper~\cite{subramanian2021torch_cka}.

\subsubsection{Dataset-based CKA Difference (\texttt{DS-diff})}

In dataset-based CKA difference (\texttt{DS-diff}), two self-similarity CKA analyses are performed, and the difference of these maps are used to calculate \texttt{DS-diff}. In particular:

\begin{align}
    &\texttt{DS-diff} = \nonumber \\ 
    &\overline{|CKA(m_x, ds_x, m_x, ds_x)- CKA(m_x, ds_y, m_x, ds_y)|},
\end{align}
\noindent
where the single model, $m_x$, had been trained over dataset $ds_x$. Furthermore, the subtraction operator between the two CKA maps yields a single map of the same dimensions as the inputs, the absolute value operation yields a single map of the same dimensions as the input consisting of only absolute values, and the mean operation yields a single scalar value. The \texttt{DS-diff} increases as the two CKA maps become less similar, which we hypothesize occurs due to a more extreme domain shift between datasets $ds_x$ and $ds_y$. Therefore, we expect a larger \texttt{DS-diff} to positively correlate with the MAE of $m_x$ evaluated over $ds_y$.

\subsubsection{Dataset-based CKA Similarity (\texttt{DS-sim})}

In dataset-based CKA similarity (\texttt{DS-sim}), a single CKA analysis is performed contrasting two datasets. In particular:

\begin{equation}
    \texttt{DS-sim} = \overline{diag(CKA(m_x, ds_x, m_x, ds_y))},
\end{equation}
\noindent
where the model $m_x$ had been trained over dataset $ds_x$, but the CKA analysis compares its behavior across both datasets $ds_x$ and $ds_y$. The diagonal of the CKA map is taken before averaging so that the analysis only takes same-layer similarities into consideration, thus eliminating some of the noise inherent with comparing dissimilar layers. This was not done for \texttt{DS-diff} due to its CKA analyses being self-similarity, which results in a diagonal of perfectly similar CKA values.

While \texttt{DS-sim} may appear plausible, the specific comparison made by CKA is on the feature level as encoded by the networks, so we realize that varying the dataset within the CKA similarity comparison may dilute the dataset-centric signal that we are targeting with incidental features resulting simply from a mismatch of data.

The above detraction against \texttt{DS-sim} notwithstanding, we expect a greater \texttt{DS-sim} value to signify that the compared datasets $ds_x$ and $ds_y$ are more similar, \ie that the domain shift between them is less severe. Therefore, we expect a larger \texttt{DS-sim} to negatively correlate with the MAE of $m_x$ evaluated over $ds_y$.

\subsubsection{Model-based CKA Similarity (\texttt{Model-sim})}

Model-based CKA similarity (\texttt{Model-sim}) functions similarly to \texttt{DS-sim} in that a single CKA analysis is performed, but rather than the analysis capturing a single model over two datasets, two models over a single dataset are compared. In particular:

\begin{equation}
    \texttt{Model-sim} = \overline{diag(CKA(m_x, ds_y, m_y, ds_y))},
\end{equation}
\noindent
where the models $m_x$ and $m_y$ had been trained over datasets $ds_x$ and $ds_y$ respectively. As was the case with \texttt{DS-sim}, the diagonal of the CKA map is taken before averaging for the purpose of noise reduction.

This metric has a severe flaw limiting its deployment: Unlike \texttt{DS-diff} and \texttt{DS-sim}, \texttt{Model-sim} requires two separate models, one trained on each of the compared datasets. However, gauging domain shifts between a known training dataset and in-the-wild data is a natural application of this work. We include this metric despite it being limited to the comparison of datasets with ground truth available.

As was the case with \texttt{DS-sim}, we expect a greater \texttt{Model-sim} value to signify that the compared datasets $ds_x$ and $ds_y$ are more similar, resulting in a negative correlation with the MAE of $m_x$ evaluated over $ds_y$.
\section{Domain Shift Metric Evaluation}

\begin{figure*}
    \centering
    \subfloat[MAE]{\includegraphics[width=.49\linewidth]{figs/xds-heatmap.pdf} \label{fig:domainshift-heatmaps:mae}}
    \subfloat[\texttt{DS-diff}]{\includegraphics[width=.49\linewidth]{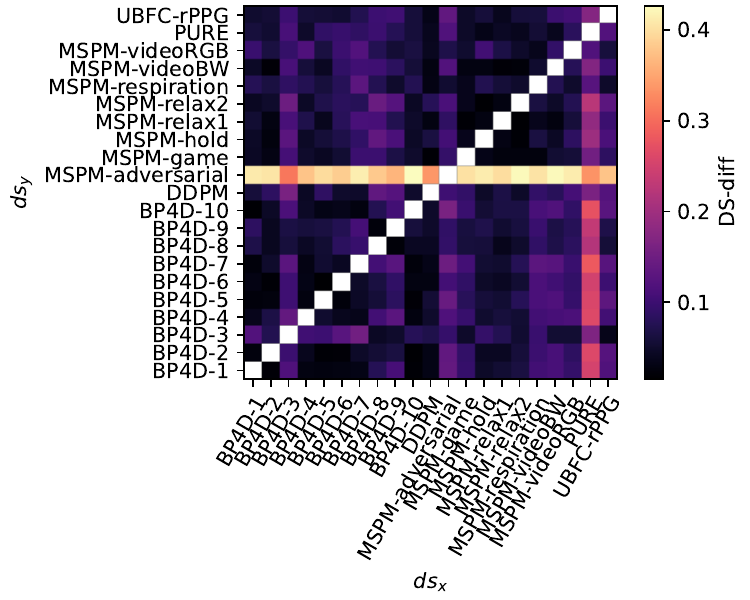} \label{fig:domainshift-heatmaps:DS-diff}} \\
    \subfloat[\texttt{DS-sim}]{\includegraphics[width=.49\linewidth]{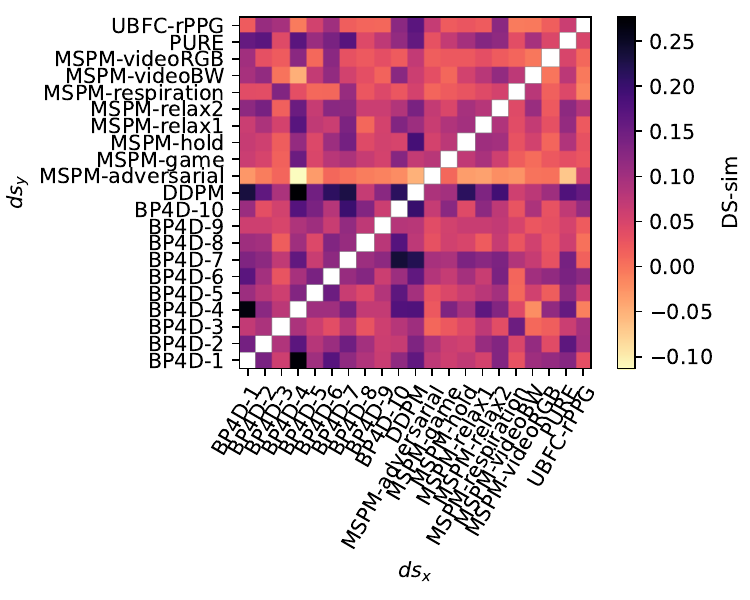} \label{fig:domainshift-heatmaps:DS-sim}}
    \subfloat[\texttt{Model-sim}]{\includegraphics[width=.49\linewidth]{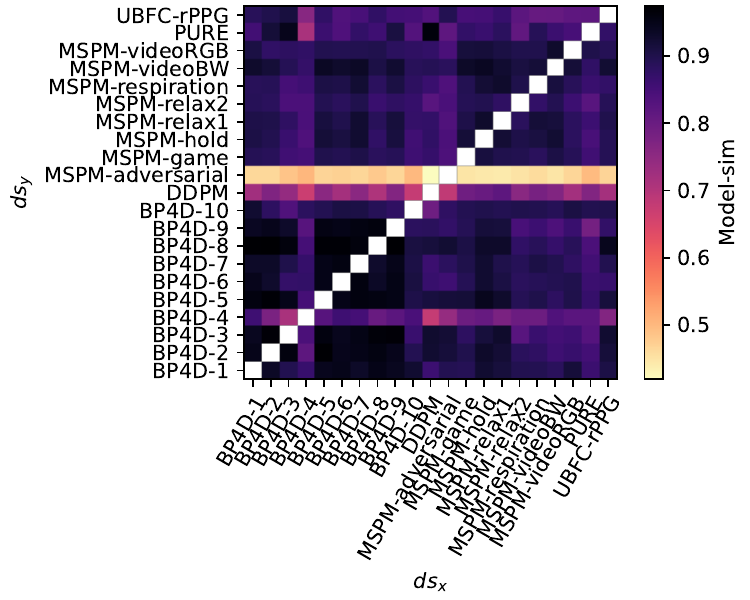} \label{fig:domainshift-heatmaps:Model-sim}}
    \caption{Heatmaps of results for the metrics over every combination of datasets averaged over the five folds. \ref{fig:domainshift-heatmaps:mae} shows the MAE as in Figure \ref{fig:domainshift_xds}, while the others show their metric for each combination of $ds_x$ and $ds_y$. Figures \ref{fig:domainshift-heatmaps:DS-sim} and \ref{fig:domainshift-heatmaps:Model-sim} use an inverted colormap for a clear comparison.}
    \label{fig:domainshift-heatmaps}
\end{figure*}

\begin{table}
    \centering
    \begin{tabular}{l|ccc}
        \toprule
        Train Dataset    & DS-diff        & DS-sim          & Model-sim       \\
        \midrule
        BP4D-1           & \textbf{0.769} & -0.041          & \textbf{-0.933} \\
        BP4D-2           & \textbf{0.857} & -0.206          & \textbf{-0.911} \\
        BP4D-3           & \textbf{0.805} & -0.103          & \textbf{-0.881} \\
        BP4D-4           & \textbf{0.715} & -0.214          & \textbf{-0.792} \\
        BP4D-5           & \textbf{0.825} & -0.197          & \textbf{-0.919} \\
        BP4D-6           & \textbf{0.719} & -0.103          & \textbf{-0.899} \\
        BP4D-7           & \textbf{0.639} & -0.052          & \textbf{-0.912} \\
        BP4D-8           & \textbf{0.816} & -0.215          & \textbf{-0.918} \\
        BP4D-9           & \textbf{0.845} & -0.191          & \textbf{-0.920} \\
        BP4D-10          & \textbf{0.758} & -0.163          & \textbf{-0.911} \\
        DDPM             & \textbf{0.936} & \textbf{-0.506} & \textbf{-0.904} \\
        MSPM-adversarial & -0.037         & \textbf{0.286}  & \textbf{-0.818} \\
        MSPM-game        & \textbf{0.889} & 0.015           & \textbf{-0.913} \\
        MSPM-hold        & \textbf{0.811} & -0.100          & \textbf{-0.910} \\
        MSPM-relax1      & \textbf{0.886} & \textbf{-0.284} & \textbf{-0.923} \\
        MSPM-relax2      & \textbf{0.824} & -0.185          & \textbf{-0.889} \\
        MSPM-respiration & \textbf{0.774} & \textbf{-0.301} & \textbf{-0.881} \\
        MSPM-videoBW     & \textbf{0.805} & -0.244          & \textbf{-0.868} \\
        MSPM-videoRGB    & \textbf{0.761} & 0.189           & \textbf{-0.868} \\
        PURE             & \textbf{0.422} & -0.240          & \textbf{-0.807} \\
        UBFC-rPPG        & \textbf{0.798} & \textbf{0.287}  & \textbf{-0.933} \\
        \midrule
        Composite        & \textbf{0.781} & -0.125          & \textbf{-0.896} \\
        \bottomrule
    \end{tabular}
    \caption{Correlation of using CKA to measure domain shift. Bolded $r$ values are significant at Bonferroni corrected $p < 0.0024$. Composite correlations are calculated using the average of the Fisher z transformed correlation coefficients~\cite{corey1998averaging}.}
    \label{tab:domainshift_ckacorrelation}
\end{table}

A useful domain shift metric should be correlated with empirical performance. Results of correlating the domain shift metrics with MAE are reported in Table \ref{tab:domainshift_ckacorrelation}. For each pair of datasets $ds_x$ and $ds_y$, the metrics as described in Section \ref{subsec:domainshift_metrics} are calculated, as well as the MAE of a model trained on $ds_x$ and evaluated on $ds_y$. Then a linear regression is performed across all datasets $ds_y$ (while holding $ds_x$ constant) calculating the correlation between the evaluated metric and MAE. These results are reported in Table \ref{tab:domainshift_ckacorrelation}.

We found that the \texttt{DS-diff} and \texttt{Model-sim} metrics performed well as CKA-based predictors of MAE, which is a proxy for domain shift. We did not find significant results for \texttt{DS-sim}. This is a result of features not being well correlated between the datasets. Consequently, the analysis is overwhelmed by non domain shift differences between the data (\eg different pulse phase) such that it does not function as a predictor of a domain shift.

An additional representation of the results for each metric is reported in Figure \ref{fig:domainshift-heatmaps}. We include the cross dataset MAE results in Figure \ref{fig:domainshift-heatmaps:mae} for the sake of comparison since we expect the CKA-based metrics to exhibit similar behavior to the empirical MAE results.

Indeed, in Figure \ref{fig:domainshift-heatmaps:DS-diff} we observe a strong \texttt{DS-diff} for all cases where $ds_y$ is MSPM-adversarial, which is in keeping with the MAE results from \ref{fig:domainshift-heatmaps:mae}. The \texttt{Model-sim} metric in Figure \ref{fig:domainshift-heatmaps:Model-sim} further exhibits similar trends regarding DDPM and BP4D-4, which are not visually apparent with \texttt{DS-diff}.

We did not expect to observe clear trends for \texttt{DS-sim} in Figure \ref{fig:domainshift-heatmaps:DS-sim} given its poor correlation with MAE as reported in Table \ref{tab:domainshift_ckacorrelation}. However, we do still observe low similarity where $ds_y$ is MSPM-adversarial, indicating that while the \texttt{DS-sim} metric is not strong, in extreme cases it can still reveal a domain shift.

We find it noteworthy that all three of the metrics could be used to predict the success of the attack on rPPG in MSPM-adversarial. This result is especially useful as the \texttt{DS-diff} and \texttt{DS-sim} metrics do not require ground truth for MSPM-adversarial in their analyses, \ie similar predictions could be made for in-the-wild data.

The most clear trends in Figure \ref{fig:domainshift-heatmaps} affect full rows of the dataset, \eg MSPM-adversarial as $ds_y$ (or the Test Dataset) constitutes a severe domain shift according to all of the evaluated metrics. However, less prominent yet still present trends exist in the columns of Figures \ref{fig:domainshift-heatmaps:DS-diff} and \ref{fig:domainshift-heatmaps:Model-sim}. These column-wise trends indicate datasets which, when used to train models, result in more or less severe domain shifts when transferring to the other compared datasets. In particular, when taking the median value for each column in \ref{fig:domainshift-heatmaps:DS-diff}, the top five $ds_x$ datasets by \texttt{DS-diff} are, in increasing order, BP4D-8, UBFC-rPPG, MSPM-adversarial, BP4D-3, and PURE. Similarly, the bottom five $ds_x$ datasets by median \texttt{Model-sim} in \ref{fig:domainshift-heatmaps:Model-sim} are, in decreasing order, BP4D-10, DDPM, MSPM-adversarial, PURE, and BP4D-4. Finally, the top five training datasets by MAE in \ref{fig:domainshift-heatmaps:mae} are, in increasing order, UBFC-rPPG, PURE, BP4D-4, BP4D-10, and MSPM-adversarial. While the agreement is not perfect, it appears that PURE and MSPM-adversarial stand out as poor training datasets for the goal of mitigating domain shift related issues when transferring to other domains.

\subsection{Application: Model Selection} \label{sec:modelselect}

A natural application of this work is model selection. In particular, using \texttt{DS-diff}, which is applicable for an unknown $ds_y$, we calculate the domain shift from all models trained on known source datasets $ds_x$, selecting the one which minimizes the metric for evaluation.

\begin{table}
    \centering
    \begin{tabular}{c|c}
        \toprule
        Case & Average MAE \\
        \midrule
        Worst & 10.301 $\pm$ 1.131 \\
        Average & 7.314 $\pm$ 0.136 \\
        Best & 5.545 $\pm$ 0.096 \\
        \texttt{DS-diff} & 6.888 $\pm$ 0.182 \\
        \bottomrule
    \end{tabular}
    \caption{Empirical errors of worst case, average case, and best case model selection compared with \texttt{DS-diff} as the method for model selection. Results are averaged over all datasets. 95\% confidence intervals are across 5 folds.}
    \label{tab:modelselection}
\end{table}

\begin{figure}
    \centering
    \includegraphics[width=\linewidth]{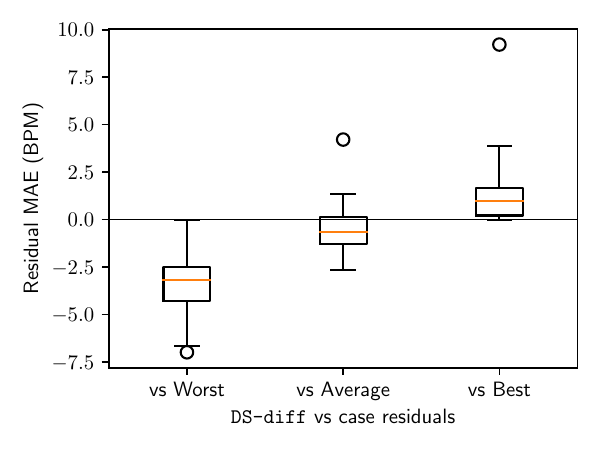}
    \caption{Residual MAE for \texttt{DS-diff} compared to each baseline, calculated over each dataset targeted as $ds_y$.}
    \label{fig:ds-diff-boxplot}
\end{figure}

Empirical results for using this scheme for model selection are presented in Table \ref{tab:modelselection}. We compare the \texttt{DS-diff} model selection scheme to three baseline: ``Worst'' in which the model resulting in the poorest performance is selected; ``Best'' in which the model optimizing performance is selected, and ``Average'' in which the average performance of all models is reported. We observe that \texttt{DS-diff} constitutes a significant improvement over the worst case scenario and a marginal improvement over the average case.

We further analyzed the per-dataset differences between \texttt{DS-diff} and the baselines.
Figure \ref{fig:ds-diff-boxplot} shows the residual MAE between \texttt{DS-diff} and each baseline, calculated in a pairwise manner between the methods on each of the 21 datasets. We observe that the median residual for \texttt{DS-diff} compared to the ``Worst'' case baseline is -3.2 BPM, compared to ``Average'' is -0.65 BPM, and compared to ``Best'' is 0.96 BPM. In terms of percent improvement over the non-residual errors this equates to, on average, a 41.2\% improvement over the worst case, a 13.9\% improvement over the average case, and a 32.1\% degradation compared to the best case. This result reveals that, while the solution outperforms the average case, progress can still be made to develop model similarity based methods for model selection.

We believe that the issues with applying \texttt{DS-diff} for model selection can be largely explained by certain behaviors of that metric which do not relate to empirical performance. Revisiting Figure \ref{fig:domainshift-heatmaps:DS-diff}, we observe column-level trends which differ from those present in the empirical results (Figure \ref{fig:domainshift-heatmaps:mae}). In particular, the empirical results show degraded performance in some cases for models trained on BP4D-4, BP4D-10, and MSPM-adversarial. However, while \texttt{DS-diff} shows increased domain shifts with MSPM-adversarial as $ds_x$, it does not exhibit clear column-wise trends for BP4D-4 and BP4D-10, instead exhibiting trends for BP4D-3 and PURE. While these trends may not be purely spurious and certainly do relate to how model activation behavior differs between these datasets, they do not correlate directly with empirical results, and therefore contribute to suboptimal performance in the model selection application.
\section{Conclusions} 

The key contributions of this research are the tools provided in the form of CKA-based metrics for measuring the domain shift between datasets. We demonstrate significant correlation with MAE (\ie an empirical measure for domain shift) for the \texttt{DS-diff} and \texttt{Model-sim} metrics, in which model activations are compared. We demonstrated that the metrics reveal how datasets provide domain shift related challenges as test datasets, including the detection of an adversarial attack on rPPG. The metrics also reveal how datasets may be well or ill suited as training datasets for the purpose of domain shift mitigation. Furthermore, we showed how \texttt{DS-diff} can be used for model selection in an application where ground truth is not known, demonstrating a 13.9\% improvement over the average case baseline of model selection. However, this metric highlights domain shift effects in terms of how model activation behavior differs between datasets, a phenomenon which does not always directly result in empirical performance. As a result, while this metric is useful for understanding domain shifts in terms of model representations of the data, more research may be necessary to develop model similarity based predictors of empirical performance.

One feature of our techniques for domain shift measurement is that they are based on model activations, \ie they are dependant on features that are affected by the selection of training parameters and model architecture. In this study we selected a single 3DCNN based architecture. However, an extension to this work should explore the robustness of the results under varied architectures.

{\small
\bibliographystyle{ieee_fullname}
\bibliography{main}

\begin{thebibliography}{10}\itemsep=-1pt

\bibitem{Bobbia2019}
Serge Bobbia, Richard Macwan, Yannick Benezeth, Alamin Mansouri, and Julien
  Dubois.
\newblock {Unsupervised skin tissue segmentation for remote
  photoplethysmography}.
\newblock {\em Pattern Recognition Letters}, 124:82--90, 2019.

\bibitem{chakrabarty2023simple}
Goirik Chakrabarty, Manogna Sreenivas, and Soma Biswas.
\newblock A simple signal for domain shift.
\newblock In {\em Proceedings of the IEEE/CVF International Conference on
  Computer Vision}, pages 3577--3584, 2023.

\bibitem{corey1998averaging}
David~M Corey, William~P Dunlap, and Michael~J Burke.
\newblock Averaging correlations: Expected values and bias in combined pearson
  rs and fisher's z transformations.
\newblock {\em The Journal of general psychology}, 125(3):245--261, 1998.

\bibitem{cui2022deconfounded}
Tianyu Cui, Yogesh Kumar, Pekka Marttinen, and Samuel Kaski.
\newblock Deconfounded representation similarity for comparison of neural
  networks.
\newblock {\em Advances in Neural Information Processing Systems},
  35:19138--19151, 2022.

\bibitem{DeHaan2013}
G. {de Haan} and V. {Jeanne}.
\newblock Robust pulse rate from chrominance-based rppg.
\newblock {\em IEEE Trans. on Biom. Eng.}, 60(10):2878--2886, 2013.

\bibitem{du2023dual}
Jingda Du, Si-Qi Liu, Bochao Zhang, and Pong~C Yuen.
\newblock Dual-bridging with adversarial noise generation for domain adaptive
  rppg estimation.
\newblock In {\em Proceedings of the IEEE/CVF Conference on Computer Vision and
  Pattern Recognition}, pages 10355--10364, 2023.

\bibitem{elsahar2019annotate}
Hady Elsahar and Matthias Gall{\'e}.
\newblock To annotate or not? predicting performance drop under domain shift.
\newblock In {\em Proceedings of the 2019 Conference on Empirical Methods in
  Natural Language Processing and the 9th International Joint Conference on
  Natural Language Processing (EMNLP-IJCNLP)}, pages 2163--2173, 2019.

\bibitem{gretton2005measuring}
Arthur Gretton, Olivier Bousquet, Alex Smola, and Bernhard Sch{\"o}lkopf.
\newblock Measuring statistical dependence with hilbert-schmidt norms.
\newblock In {\em International conference on algorithmic learning theory},
  pages 63--77. Springer, 2005.

\bibitem{kanakasabapathy2021adaptive}
Manoj~Kumar Kanakasabapathy, Prudhvi Thirumalaraju, Hemanth Kandula, Fenil
  Doshi, Anjali~Devi Sivakumar, Deeksha Kartik, Raghav Gupta, Rohan Pooniwala,
  John~A Branda, Athe~M Tsibris, et~al.
\newblock Adaptive adversarial neural networks for the analysis of lossy and
  domain-shifted datasets of medical images.
\newblock {\em Nature biomedical engineering}, 5(6):571--585, 2021.

\bibitem{koenecke2020racial}
Allison Koenecke, Andrew Nam, Emily Lake, Joe Nudell, Minnie Quartey, Zion
  Mengesha, Connor Toups, John~R Rickford, Dan Jurafsky, and Sharad Goel.
\newblock Racial disparities in automated speech recognition.
\newblock {\em Proceedings of the National Academy of Sciences},
  117(14):7684--7689, 2020.

\bibitem{kornblith2019similarity}
Simon Kornblith, Mohammad Norouzi, Honglak Lee, and Geoffrey Hinton.
\newblock Similarity of neural network representations revisited.
\newblock In {\em International conference on machine learning}, pages
  3519--3529. PMLR, 2019.

\bibitem{laakso2000content}
Aarre Laakso and Garrison Cottrell.
\newblock Content and cluster analysis: assessing representational similarity
  in neural systems.
\newblock {\em Philosophical psychology}, 13(1):47--76, 2000.

\bibitem{liu2020multi}
Xin Liu, Josh Fromm, Shwetak Patel, and Daniel McDuff.
\newblock Multi-task temporal shift attention networks for on-device
  contactless vitals measurement.
\newblock {\em Advances in Neural Information Processing Systems},
  33:19400--19411, 2020.

\bibitem{luo2019taking}
Yawei Luo, Liang Zheng, Tao Guan, Junqing Yu, and Yi Yang.
\newblock Taking a closer look at domain shift: Category-level adversaries for
  semantics consistent domain adaptation.
\newblock In {\em Proceedings of the IEEE/CVF conference on computer vision and
  pattern recognition}, pages 2507--2516, 2019.

\bibitem{mengesha2021don}
Zion Mengesha, Courtney Heldreth, Michal Lahav, Juliana Sublewski, and Elyse
  Tuennerman.
\newblock “i don’t think these devices are very culturally
  sensitive.”—impact of automated speech recognition errors on african
  americans.
\newblock {\em Frontiers in Artificial Intelligence}, 4:169, 2021.

\bibitem{morcos2018insights}
Ari Morcos, Maithra Raghu, and Samy Bengio.
\newblock Insights on representational similarity in neural networks with
  canonical correlation.
\newblock {\em Advances in neural information processing systems}, 31, 2018.

\bibitem{raghu2017svcca}
Maithra Raghu, Justin Gilmer, Jason Yosinski, and Jascha Sohl-Dickstein.
\newblock Svcca: Singular vector canonical correlation analysis for deep
  learning dynamics and interpretability.
\newblock {\em Advances in neural information processing systems}, 30, 2017.

\bibitem{sankaranarayanan2018learning}
Swami Sankaranarayanan, Yogesh Balaji, Arpit Jain, Ser~Nam Lim, and Rama
  Chellappa.
\newblock Learning from synthetic data: Addressing domain shift for semantic
  segmentation.
\newblock In {\em Proceedings of the IEEE conference on computer vision and
  pattern recognition}, pages 3752--3761, 2018.

\bibitem{speth2024mspm}
Jeremy Speth, Nathan Vance, Benjamin Sporrer, Lu Niu, Patrick Flynn, and Adam
  Czajka.
\newblock Mspm: A multi-site physiological monitoring dataset for remote pulse,
  respiration, and blood pressure estimation.
\newblock {\em arXiv preprint arXiv:2402.02224}, 2024.

\bibitem{stacke2020measuring}
Karin Stacke, Gabriel Eilertsen, Jonas Unger, and Claes Lundstr{\"o}m.
\newblock Measuring domain shift for deep learning in histopathology.
\newblock {\em IEEE journal of biomedical and health informatics},
  25(2):325--336, 2020.

\bibitem{Stricker2014}
Ronny Stricker, Steffen Muller, and Horst~Michael Gross.
\newblock {Non-contact video-based pulse rate measurement on a mobile service
  robot}.
\newblock {\em IEEE International Symposium on Robot and Human Interactive
  Communication}, pages 1056--1062, 2014.

\bibitem{subramanian2021torch_cka}
Anand Subramanian.
\newblock torch\_cka, 2021.

\bibitem{tang2023mmpd}
Jiankai Tang, Kequan Chen, Yuntao Wang, Yuanchun Shi, Shwetak Patel, Daniel
  McDuff, and Xin Liu.
\newblock Mmpd: Multi-domain mobile video physiology dataset.
\newblock {\em arXiv preprint arXiv:2302.03840}, 2023.

\bibitem{vance2024cka}
Nathan Vance and Patrick Flynn.
\newblock Refining remote photoplethysmography architectures using cka and
  empirical methods.
\newblock {\em arXiv preprint arXiv:2401.04801}, 2024.

\bibitem{vance2022deception}
Nathan Vance, Jeremy Speth, Siamul Khan, Adam Czajka, Kevin~W. Bowyer, Diane
  Wright, and Patrick Flynn.
\newblock Deception detection and remote physiological monitoring: A dataset
  and baseline experimental results.
\newblock {\em IEEE Transactions on Biometrics, Behavior, and Identity Science
  (TBIOM)}, pages 1--1, 2022.

\bibitem{vance2023generalization}
Nathan Vance, Jeremy Speth, Benjamin Sporrer, and Patrick Flynn.
\newblock Promoting generalization in cross-dataset remote
  photoplethysmography.
\newblock In {\em Proceedings of the IEEE/CVF Conference on Computer Vision and
  Pattern Recognition (CVPR) Workshops}, pages 5984--5992, 2023.

\bibitem{wang2023hierarchical}
Jiyao Wang, Hao Lu, Ange Wang, Yingcong Chen, and Dengbo He.
\newblock Hierarchical style-aware domain generalization for remote
  physiological measurement.
\newblock {\em IEEE Journal of Biomedical and Health Informatics}, 2023.

\bibitem{wang2016algorithmic}
Wenjin Wang, Albertus~C Den~Brinker, Sander Stuijk, and Gerard De~Haan.
\newblock Algorithmic principles of remote ppg.
\newblock {\em IEEE Transactions on Biomedical Engineering}, 64(7):1479--1491,
  2016.

\bibitem{yu2019remote}
Zitong Yu, Xiaobai Li, and Guoying Zhao.
\newblock Remote photoplethysmograph signal measurement from facial videos
  using spatio-temporal networks.
\newblock {\em arXiv preprint arXiv:1905.02419}, 2019.

\bibitem{zhang2021adaptive}
Marvin Zhang, Henrik Marklund, Nikita Dhawan, Abhishek Gupta, Sergey Levine,
  and Chelsea Finn.
\newblock Adaptive risk minimization: Learning to adapt to domain shift.
\newblock {\em Advances in Neural Information Processing Systems},
  34:23664--23678, 2021.

\bibitem{zhang2016multimodal}
Zheng Zhang, Jeff~M Girard, Yue Wu, Xing Zhang, Peng Liu, Umur Ciftci, Shaun
  Canavan, Michael Reale, Andy Horowitz, Huiyuan Yang, et~al.
\newblock Multimodal spontaneous emotion corpus for human behavior analysis.
\newblock In {\em Proceedings of the IEEE conference on computer vision and
  pattern recognition}, pages 3438--3446, 2016.

\end{thebibliography}
}


\end{document}